\documentclass{article}

\usepackage[utf8]{inputenc}
\usepackage[T1]{fontenc}
\usepackage[english]{babel}
\usepackage[nottoc]{tocbibind}
\usepackage{graphicx}
\usepackage{amsmath}
\usepackage{algorithm2e}
\usepackage{float}
\usepackage{siunitx}
\usepackage{appendix}

\graphicspath{ {./images/} }

\title{Neural Network Processing Neural Networks: An efficient way to learn higher order functions}
\author{Firat Tuna\\
  \texttt{firat.tuna@kcl.ac.uk}
}
\date{June 2019}

\begin{document}

\maketitle
\begin{abstract}
Functions are rich in meaning and can be interpreted in a variety of ways. Neural networks were proven to be capable of approximating a large class of functions\cite{universalapproximator}. In this paper, we propose  a new class of neural networks called "Neural Network Processing Neural Networks" (NNPNNs), which inputs neural networks and numerical values, instead of just numerical values. Thus enabling neural networks to represent and process rich structures.
\end{abstract}

\section{Introduction}
 Continuous functions can be interpreted to represent many things from probability distributions to graphics, yet neural networks, despite proven to be effective in reasoning with other entities, can not reason with them effectively.
\smallskip

\section{Model Overview}
NNPNNs are neural networks which include queries to an inputted neural network $G$, in effect, it searches the neural network by trying on different inputs and processing the input-output pairs. 
\\
\\
To do that NNPNNs consists of $l$ super-layers which we call 'phases', which each use a combination of dense layers (a processing block) to generate $r$ inputs ($x_n$) to $G$. The outputs of each read $G(x_n)$ are then concatenated with the inputs to get ($G(x_n), x_n$) which is inputted to the next phase. The outputs of the last phase are inputted to another processing block to generate the output of the NNPNN.

\begin{figure}[H]
    \noindent\makebox[\textwidth]{\includegraphics[scale=0.3]{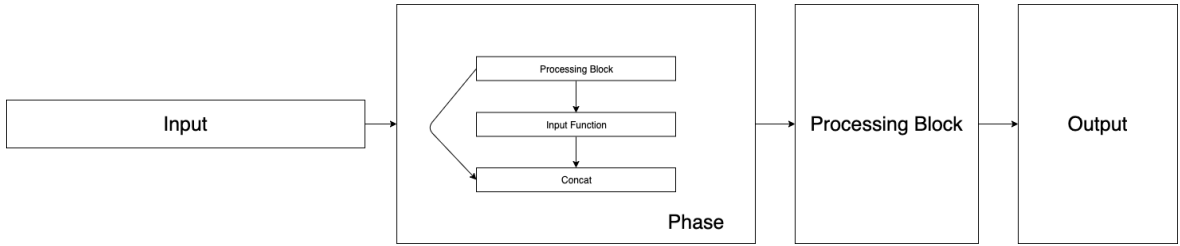}}
    \caption{an NNPNN with $l=1$}
    \label{fig:my_label}
\end{figure}

\section{Experiments}
This section describes the experiments conducted to test the effectiveness of NNPNNs in processing neural networks. The goal was to show that NNPNNs can solve significant problems while being fast and differentiable.
\\
\\
For the experiments 1 and 2, as there were no alternatives which were fast and differentiable, we checked if the error rate for unseen input decreased substantially, and took that fact to mean that the model was capable of learning the objective higher-order function. 
\\
\\
For the experiment 3, which didn't approximate a higher order function, we compared our network's performance with that of a residual network's. 
\subsection{General Inverse Function}
\subsubsection{Definition}
General Inverse function is a function which can approximate any inverse function. An inverse function is one with the property:
$$f(x) = y \Leftrightarrow f^{-1}(y) = x$$
Hence, we aim to train the neural network G where:
$$G(F,x) = F^{-1}(x) \Leftrightarrow G(F,F(x)) = x$$
\medskip
We train such an NNPNN with the following procedure:
\smallbreak
\begin{algorithm}[H]
    \While{training}{
        $G\gets generate\_NN() $\footnote{$generate\_NN()$ generates a neural network with randomly initialized weights and number of hidden layers from 1 to 5 with 5 hidden units each.}\;
        $G_{input}\gets random\_input() $\footnote{$random\_input()$ is drawn from a Gaussian distribution with the mean of 0.0 and variance of 100.0}\;
        $F_{input}\gets G(G_{input}) $\;
        $F_{output}\gets F(F_{input}, G) $\;
        $train(F_{input},G(F_{output}))$\footnote{$train(x,y)$ optimizes the MAE Loss between x and y.}\;
    }
\end{algorithm}

\subsubsection{Results}
The resulting network was able to get outputs that were within the $10\%$ of the Manhattan Norm $54.1\%$ of the time and within the $25\%$, $75.8\%$ of the time. The average amount of such deviation was $21.6\%$, while the median amount was $8.6\%$.
\begin{figure}[H]
    \noindent\makebox[\textwidth]{\includegraphics[scale=0.5]{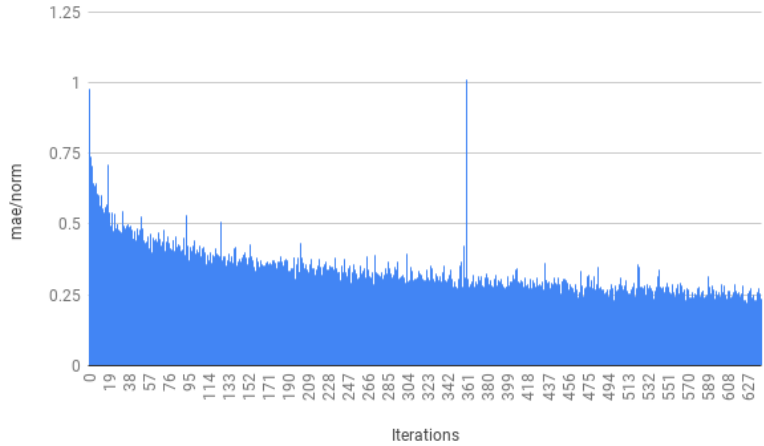}}
    \caption{$\frac{Mean Absolute Error}{Manhattan Norm}$ per number of iterations during training}
    \label{fig:norm}
\end{figure}

\begin{figure}[H]
    \noindent\makebox[\textwidth]{\includegraphics[scale=0.5]{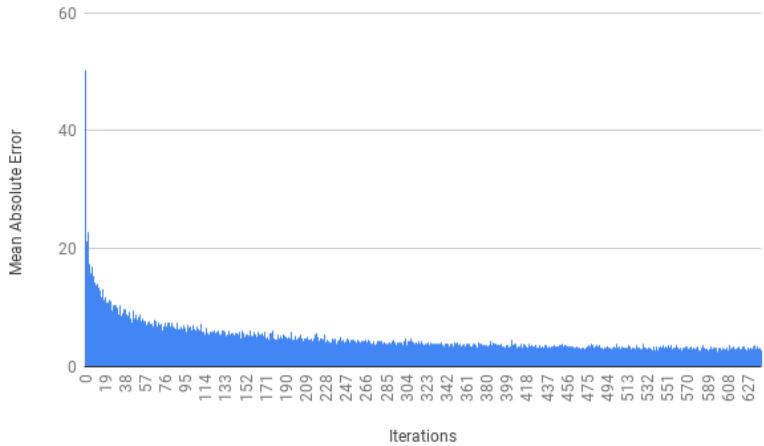}}
    \caption{Mean Absolute Error (sum of 2 output units) per number of iterations during training}
    \label{fig:mae}
\end{figure}

The loss, measured throughout training, consistently decreased and at the end, the model was able to make useful predictions.

\subsection{Compression}\label{exp:synthesis}
We aim to create a neural network that expresses any neural network with lesser number of parameters.
\smallbreak
To achieve that we use a meta-parameterized neural network, $F_2$, which takes the output of NNPNN $F_1$ and any $x$ and outputs $G(x)$.
\smallbreak
Thus, we aim to train $F_1$, $F_2$ approximating the following condition:
$$F_1(G)=x_{meta} \implies F_2(x,x_{meta}) = G(x)$$

where the size of $x_{meta}$ is lesser than the number of parameters for $G(x)$, thus constituting a compression.
\smallbreak
To train 
We use the following procedure to train $F_1, F_2$:
\smallbreak
\begin{algorithm}[H]
    \While{training}{
        $G\gets generate\_NN() $\footnote{$generate\_NN()$ generates a neural network with randomly initialized weights and number of hidden layers from 1 to 5 with 5 hidden units each.}\;
        $x_{meta}\gets F_1(G) $\;
        $x\gets random\_input() $\footnote{$random\_input()$ is drawn from a Gaussian distribution with the mean of 0.0 and variance of 100.0}\;
        $train(F_2(x,x_{meta}),G(x))$\footnote{$train(x,y)$ optimizes the MSE Loss between x and y.}\;
    }
\end{algorithm}

\subsubsection{Results}
\begin{figure}[H]
    \noindent\makebox[\textwidth]{\includegraphics[scale=0.5]{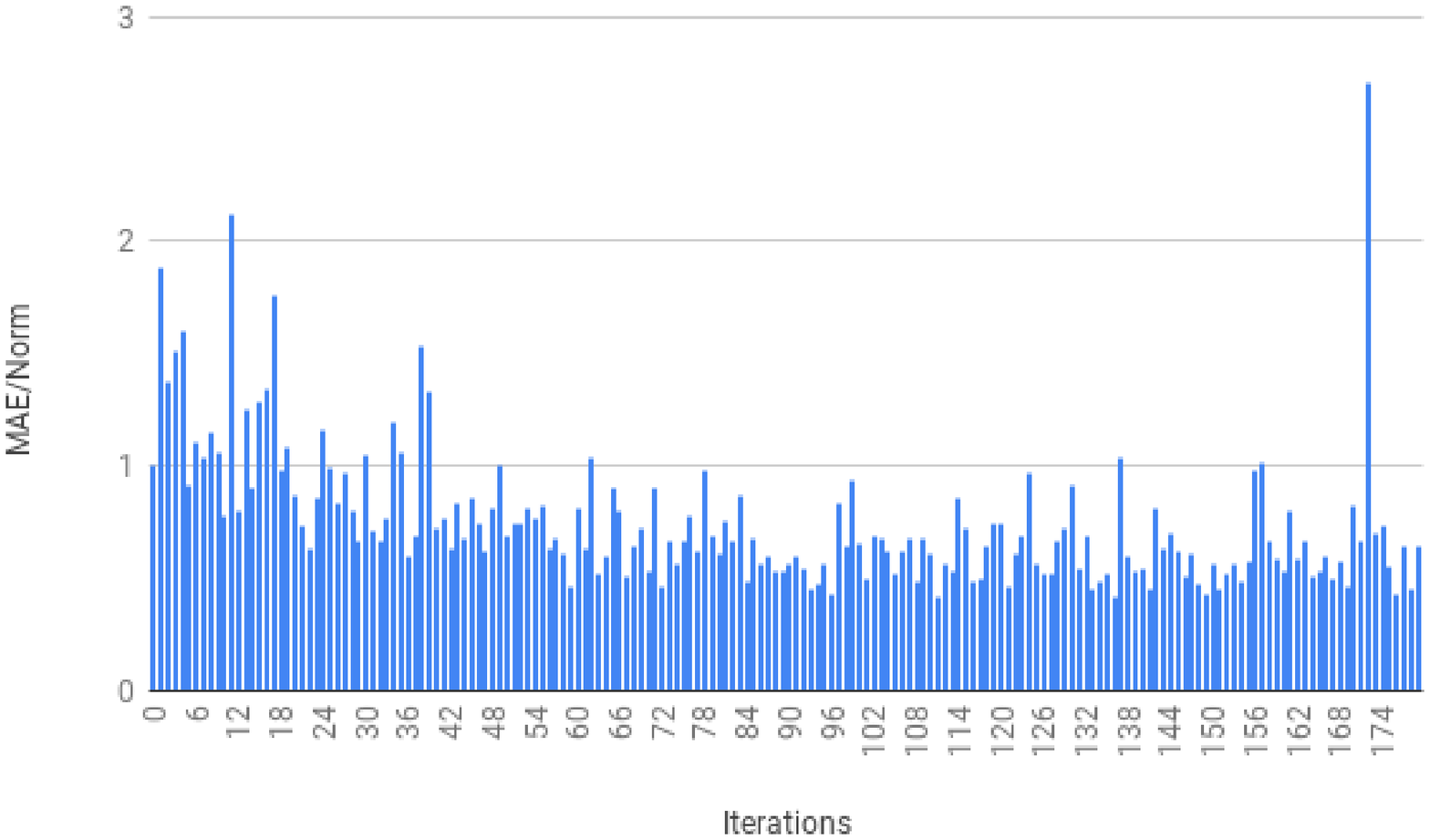}}
    \caption{$\frac{Mean Absolute Error}{Manhattan Norm}$ per number of iterations during training}
    \label{fig:syn_norm}
\end{figure}

\begin{figure}[H]
    \noindent\makebox[\textwidth]{\includegraphics[scale=0.5]{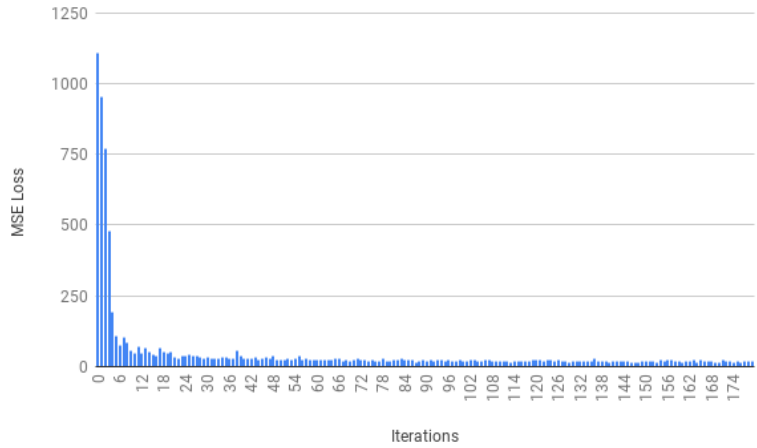}}
    \caption{Mean Squared Error (average of 2 output units) per number of iterations during training}
    \label{fig:syn_mae}
\end{figure}

As can be seen from the figures \ref{fig:syn_norm},\ref{fig:syn_mae}, the loss consistently decreased throughout the training. After the training, the model was tested on 100,000 randomly generated examples and a median loss\footnote{Mean Squared Error} of 3.8 and a mean loss of 19.2 was recorded.

\subsection{Object search}\label{exp:objloc}
is to find the coordinates for the corners of the smallest box of the biggest object in an image. For this, we have compared two functions Object
\smallbreak
As NNPNN's require neural network as an input, we make a simple ResNet output a meta parameterized neural network, feeding that neural network to the NNPNN. Both neural networks had approximately 30M parameters.
\smallbreak
\subsubsection{Results}
\begin{figure}[H]
    \noindent\makebox[\textwidth]{\includegraphics[scale=0.5]{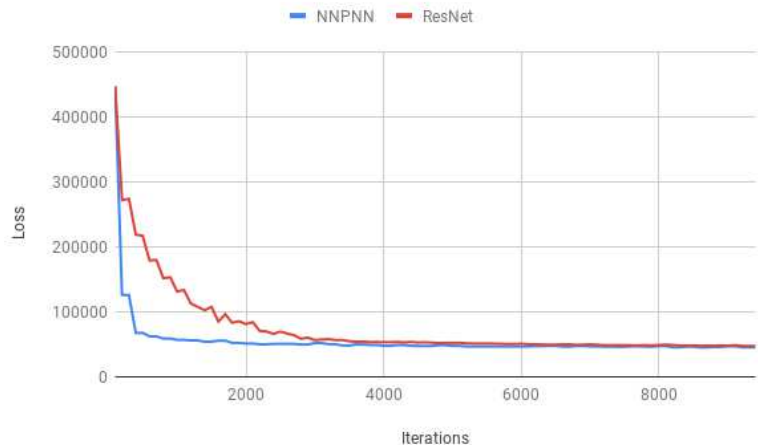}}
    \caption{Loss functions for ResNet and NNPNN}
    \label{fig:search:norm}
\end{figure}
As can be seen from the chart, NNPNN learned to do the task much faster.

\subsection{Experiment Details}
All the experiments are done with RMSProb algorithm with the learning rate of \num{2e-5}.
\section{Conclusion}
We have shown that neural networks of the described model can be trained to solve problems pertaining to neural networks. They can search and process neural networks and answer meaningful questions.
\smallbreak
We think such models might be used as a building block for a variety of artificial intelligence problems.
\smallbreak

\section{Acknowledgements}
This research did not receive any specific grant from funding agencies in the public, commercial, or not-for-profit sectors.

\appendix
\section{Experiment details}
Throughout training, RMSProp\cite{rmsprop} was used without exceptions.
\subsection{Processing Block}
A processing block is a combination of dense layers, we have used a sub-block which constitutes of 3 dense layers, first two respectively processing the input to the sub-block and the output from the first layer and the latest processing the input and the outputs from the first and second layers concatenated. In each of the following experiments, we have used 2 sub-blocks for each processing block.
\subsection{Meta parameterized neural network}
There are many effective ways to generate neural networks by other neural networks\cite{hypernetwork}, but as the purpose of this article is mainly using neural networks as inputs, we have invented a simple one called meta parameterized neural network. A meta parameterized network as referred above means a neural network which takes two sets of parameters, inputs and meta parameters, the latter intended to define a neural network, hence making it possible to generate neural networks. For our experiments we have concatenated meta parameters with the neural network input and outputs of each layer, to get the inputs to each layer.

\end{document}